\documentclass{article}

\usepackage{PRIMEarxiv}

\usepackage[utf8]{inputenc} 
\usepackage[T1]{fontenc}    
\usepackage{hyperref}       
\usepackage{url}            
\usepackage{booktabs}       
\usepackage{amsfonts}       
\usepackage{nicefrac}       
\usepackage{microtype}      
\usepackage{lipsum}
\usepackage{fancyhdr}       
\usepackage{amsmath}
\usepackage{graphicx}       
\usepackage{algorithm}
\usepackage{algpseudocode}
\usepackage{tikz}
\usetikzlibrary{shapes.geometric, arrows.meta, positioning, calc}
\usepackage{caption}
\graphicspath{{media/}}     

\title{VM-BeautyNet: A Synergistic Ensemble of Vision Transformer and Mamba for Facial Beauty Prediction
}

\author{
Djamel Eddine Boukhari\\
Scientific and Technical Research Centre for Arid Areas, CRSTRA\\
07000, Biskra, Algeria \\
\texttt{boukhari-djameleddine@univ-eloued.dz} \\
}

\begin{document}
\maketitle

\begin{abstract}
Facial Beauty Prediction (FBP) is a complex and challenging computer vision task, aiming to model the subjective and intricate nature of human aesthetic perception. While deep learning models, particularly Convolutional Neural Networks (CNNs), have made significant strides, they often struggle to capture the global, holistic facial features that are critical to human judgment. Vision Transformers (ViT) address this by effectively modeling long-range spatial relationships, but their quadratic complexity can be a bottleneck. This paper introduces a novel, heterogeneous ensemble architecture, \textbf{VM-BeautyNet}, that synergistically fuses the complementary strengths of a Vision Transformer and a Mamba-based Vision model, a recent advancement in State-Space Models (SSMs). The ViT backbone excels at capturing global facial structure and symmetry, while the Mamba backbone efficiently models long-range dependencies with linear complexity, focusing on sequential features and textures. We evaluate our approach on the benchmark SCUT-FBP5500 dataset. Our proposed VM-BeautyNet achieves state-of-the-art performance, with a \textbf{Pearson Correlation (PC) of 0.9212}, a \textbf{Mean Absolute Error (MAE) of 0.2085}, and a \textbf{Root Mean Square Error (RMSE) of 0.2698}. Furthermore, through Grad-CAM visualizations, we provide interpretability analysis that confirms the complementary feature extraction of the two backbones, offering new insights into the model's decision-making process and presenting a powerful new architectural paradigm for computational aesthetics.
\end{abstract}

\keywords{Facial Beauty Prediction\and Vision Transformer (ViT)\and Mamba\and State-Space Models (SSM)\and Ensemble Learning\and Computational Aesthetics \and Model Interpretability\and Deep Learning.}

\section{Introduction}
\label{sec:intro}

The automated assessment of facial attractiveness, a task formally known as Facial Beauty Prediction (FBP), represents a significant and long-standing challenge within the field of affective computing and computer vision\cite{b1}. Human perception of beauty is inherently subjective and multifaceted, amalgamating a complex interplay of geometric proportions, textural qualities, and socio-cultural factors \cite{b2}. Developing computational models that can accurately mirror this nuanced human judgment has profound implications for a range of applications, from personalized content recommendation and digital entertainment to medical aesthetics and surgical planning\cite{b3}.

The historical trajectory of FBP research began with methods predicated on handcrafted features, which sought to encode classical aesthetic principles such as the golden ratio and facial symmetry \cite{b4}. While foundational, these approaches were often rigid and failed to capture the subtle, high-dimensional features that underlie human perception \cite{b5}. The advent of deep learning, particularly Convolutional Neural Networks (CNNs)\cite{b6}, marked a paradigm shift. Architectures like ResNet \cite{b7} demonstrated remarkable success by automatically learning hierarchical feature representations directly from data. However, the core inductive biases of CNNs locality and spatial invariance which are advantageous for object recognition, impose limitations on FBP \cite{b8}. The constrained receptive fields of convolutional operators make it challenging to explicitly model the long-range dependencies and holistic configurations that are critical for judging facial harmony and overall structural coherence\cite{b9}.

To address this limitation, the Vision Transformer (ViT) \cite{b10} has emerged as a powerful alternative. By decomposing an image into a sequence of patches and employing the self-attention mechanism, ViTs can model the global context of an image, capturing dependencies between distant facial features, such as the relationship between eye separation and jawline structure\cite{b11, b12}. This capability is theoretically ideal for FBP. Nevertheless, the efficacy of ViT comes at a significant computational cost, as the self-attention mechanism exhibits a quadratic complexity, $O(N^2)$, with respect to the number of image patches $N$, posing challenges for scalability and efficiency.

Recently, State-Space Models (SSMs), and specifically the Mamba architecture \cite{b13}, have garnered considerable attention as a highly promising alternative to Transformers for sequence modeling. Mamba's key innovation is a Selective Scan Mechanism (SSM) that allows it to modulate its recurrence based on the input content, enabling it to effectively capture long-range dependencies while maintaining a linear computational complexity, $O(N)$. The successful adaptation of this architecture for visual tasks, as demonstrated by Vision Mamba (Vim) \cite{b14}, suggests its potential for vision applications that require both efficiency and long-range modeling.

In this work, we posit that ViT and Vision Mamba possess fundamentally complementary feature extraction capabilities. We hypothesize that the ViT backbone is adept at capturing the \textit{global spatial configuration} of a face (e.g., symmetry and proportion), while the Mamba backbone is uniquely suited for efficiently modeling \textit{fine-grained sequential information} inherent in local features and textures (e.g., skin quality and contour details). To exploit this synergy, we introduce \textbf{VM-BeautyNet}, a novel heterogeneous ensemble architecture for facial beauty prediction. Our model consists of two parallel backbones, a ViT and a Vision Mamba, whose individual predictions are intelligently fused by a lightweight, learnable module. This approach allows the model to form a more comprehensive and robust assessment by integrating both holistic and detailed facial attributes. Our principal contributions are:
\begin{enumerate}
    \item We propose a novel, synergistic ensemble architecture that, to our knowledge, is the first to combine a Vision Transformer and a Mamba-based State-Space Model for the task of facial beauty prediction.
    \item We demonstrate through extensive experimentation on the benchmark SCUT-FBP5500 dataset that our VM-BeautyNet achieves state-of-the-art performance, yielding a Pearson Correlation (PC) of 0.9212, a Mean Absolute Error (MAE) of 0.2085, and a Root Mean Square Error (RMSE) of 0.2698.
    \item We provide a qualitative analysis using Gradient-weighted Class Activation Mapping (Grad-CAM) to offer insights into the model's decision-making process. These visualizations confirm the complementary nature of the two backbones, revealing their distinct areas of focus and providing interpretability to our model's superior performance.
\end{enumerate}

The remainder of this paper is organized as follows. Section 2 reviews related work. Section 3 details our proposed methodology. Section 4 presents our experimental setup and results, and Section 5 offers analysis and discussion. Finally, Section 6 concludes the paper.

\section{Related Work}
\label{sec:related_work}

The literature on automated Facial Beauty Prediction (FBP) has evolved in tandem with advancements in computer vision and machine learning. This section reviews the key architectural paradigms that have been applied to this problem, contextualizing our work within the broader research landscape.

\subsection{Convolutional Neural Networks for FBP}
The application of deep learning to FBP was primarily driven by the success of Convolutional Neural Networks (CNNs). Early works demonstrated that fine-tuning pre-trained models like ResNet \cite{b7} on FBP datasets could significantly outperform methods based on handcrafted geometric features. These models excel at learning a rich hierarchy of local features, from simple edges and textures in early layers to more complex facial components like eyes and mouths in deeper layers. Several studies have proposed modifications to standard CNN architectures to better suit the FBP task \cite{b15} , such as multi-task learning frameworks that jointly predict attractiveness and other facial attributes, or attention mechanisms designed to weight the importance of different facial regions. However, a fundamental limitation of CNNs lies in their fixed, local receptive fields \cite{b16} . This inherent architectural property makes it difficult for them to explicitly capture the long-range spatial relationships that are crucial for perceiving holistic aesthetic concepts like facial harmony and proportion, which depend on the global arrangement of features.

\subsection{Vision Transformers in Computational Aesthetics}
To overcome the locality constraints of CNNs, the Vision Transformer (ViT) \cite{b10} was introduced, adapting the highly successful Transformer architecture from natural language processing to the vision domain. By treating an image as a sequence of patches and applying a global self-attention mechanism, ViT can model dependencies between any two patches in the image, regardless of their spatial distance. This capability is exceptionally well-suited to FBP, as it allows the model to analyze, for instance, the relationship between facial width and mouth size simultaneously. Indeed, recent work such as TransFBP \cite{b17} has demonstrated the superiority of Transformer-based models over CNNs for beauty prediction, confirming that explicitly modeling global context is advantageous. Variants like DeiT \cite{b18} have further improved the data efficiency of ViTs. Despite their impressive performance, the quadratic computational complexity of the self-attention mechanism, $O(N^2)$, where $N$ is the sequence length (number of patches), remains a significant drawback, limiting its application to higher-resolution images and increasing computational overhead.

\subsection{State-Space Models and Vision Mamba}
A recent and promising line of research has explored State-Space Models (SSMs) as an alternative to Transformers for long-sequence modeling. Foundational works on structured SSMs, such as the S4 model \cite{b13}, showed that these architectures could be highly effective for tasks involving long-range dependencies. The Mamba architecture represents a significant breakthrough in this domain. Mamba introduces a \textit{Selective Scan Mechanism} (SSM) which allows the model to selectively propagate or forget information based on the input content. This endows it with the ability to model complex dependencies in a content-aware manner while maintaining linear-time complexity, $O(N)$.

The adaptation of Mamba to the vision domain, namely Vision Mamba (Vim) \cite{b14}, has demonstrated its potential as a strong and efficient vision backbone. Vim replaces the self-attention blocks of ViT with bidirectional Mamba blocks, achieving Transformer-level performance on benchmark tasks like image classification with substantially lower computational cost. To the best of our knowledge, the potential of Vision Mamba for complex regression tasks like FBP remains unexplored.

\subsection{Ensemble and Hybrid Models}
Ensemble learning is a well-established technique for improving model performance by combining predictions from multiple models. In FBP, some works have explored ensembles of different CNN architectures or hybrid models that combine CNNs with graph neural networks \cite{b19} to model relationships between facial landmarks\cite{b20}. These approaches aim to leverage diverse feature representations to form a more robust prediction. Our work extends this principle to a novel, heterogeneous pairing of architectural paradigms. We propose that the global, configuration-aware features from ViT and the efficient, sequential texture-aware features from Vision Mamba are fundamentally complementary. By creating a hybrid architecture, we aim to harness the distinct strengths of both models, which has not been previously investigated for the FBP problem.

\section{Proposed Methodology}
\label{sec:methodology}

\begin{figure*}[!ht]
\centering
\begin{tikzpicture}[
    node distance=1.5cm and 1cm,
    block/.style={rectangle, draw, fill=blue!10, text width=10em, text centered, rounded corners, minimum height=3.5em},
    branch/.style={rectangle, draw, fill=green!10, text width=9em, text centered, rounded corners, minimum height=3em},
    connector/.style={-Straight Barb, thick},
    io/.style={ellipse, draw, fill=gray!20, minimum height=3em},
    fusion/.style={circle, draw, fill=red!20, minimum size=1cm, inner sep=0pt}
]
    \node[io] (input) {Input Image ($I$)};
    \node[block, below left=0.8cm and -1cm of input] (patch_embed_vit) {Patch Embedding};
    \node[branch, below=of patch_embed_vit] (vit_backbone) {ViT Backbone\\(Self-Attention Blocks)};
    \node[block, below right=0.8cm and 1cm of input] (patch_embed_mamba) {Patch Embedding};
    \node[branch, below=of patch_embed_mamba] (mamba_backbone) {Mamba Backbone\\(Mamba Blocks)};

    \node[block, below=1cm of vit_backbone] (head_vit) {Regression Head (Linear)};
    \node[block, below=1cm of mamba_backbone] (head_mamba) {Regression Head (Linear)};
    
    \node[fusion, below=1.8cm of $(vit_backbone)!0.5!(mamba_backbone)$] (fusion_point) {$\bigoplus$};
    
    \node[block, below=0.5cm of fusion_point] (fusion_module) {Fusion Module\\($\mathcal{F}$)};
    \node[io, below=0.5cm of fusion_module] (output) {Predicted Score ($\hat{y}$)};

    \draw[connector] (input.south) -- ++(0,-0.4) -| (patch_embed_vit.north);
    \draw[connector] (input.south) -- ++(0,-0.4) -| (patch_embed_mamba.north);
    
    \draw[connector] (patch_embed_vit) -- (vit_backbone);
    \draw[connector] (patch_embed_mamba) -- (mamba_backbone);
    
    \draw[connector] (vit_backbone.south) -- ++(0,-0.5) node[midway, right] {$p_{vit}$} -- (fusion_point.west);
    \draw[connector] (mamba_backbone.south) -- ++(0,-0.5) node[midway, left] {$p_{mamba}$} -- (fusion_point.east);

    \draw[connector] (fusion_point) -- (fusion_module);
    \draw[connector] (fusion_module) -- (output);
    
    \node[text width=15em, above right=0.5cm and -1.8cm of vit_backbone.west, text=blue!60!black] {\textbf{Branch A: Global Feature Extraction}};
    \node[text width=15em, above left=0.5cm and -1.8cm of mamba_backbone.east, text=green!50!black] {\textbf{Branch B: Sequential Feature Modeling}};
\end{tikzpicture}
\caption{The overall architecture of our proposed VM-BeautyNet. An input image is processed in parallel by two distinct backbones. Branch A (ViT) captures global spatial relationships, while Branch B (Vision Mamba) models sequential features efficiently. The intermediate predictions, $p_{vit}$ and $p_{mamba}$, are then intelligently combined by the Fusion Module $\mathcal{F}$ to produce the final beauty score $\hat{y}$.}
\label{fig:architecture}
\end{figure*}
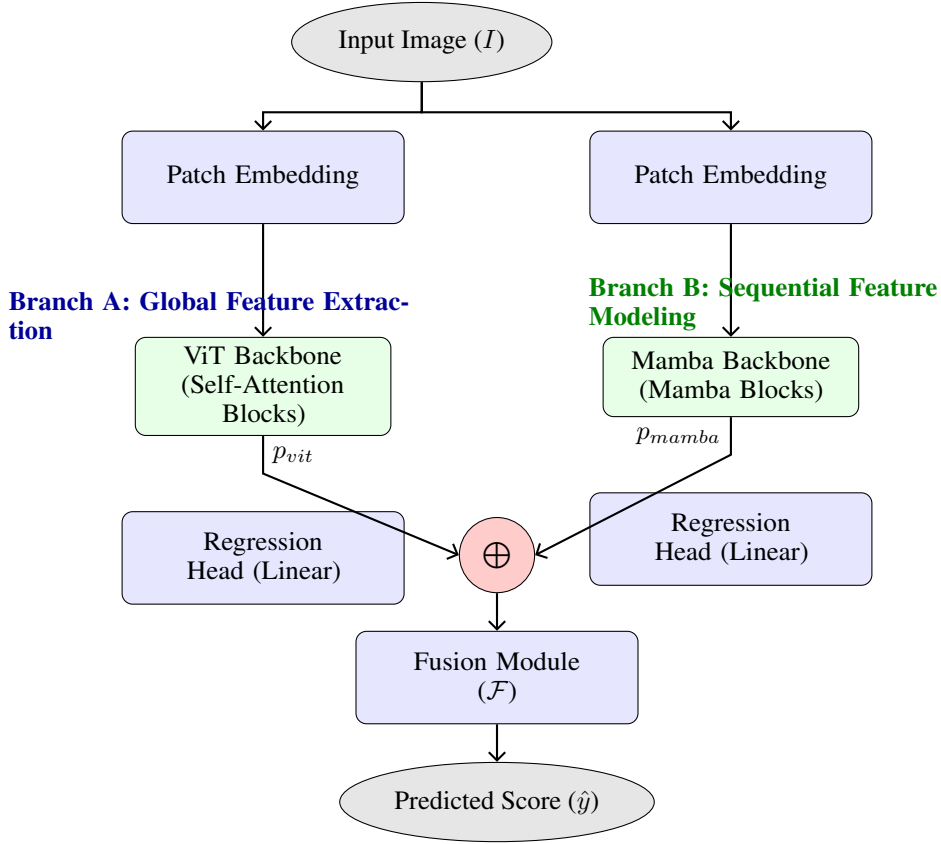

To address the multifaceted challenge of Facial Beauty Prediction (FBP), we introduce \textbf{VM-BeautyNet}, a novel dual-branch, heterogeneous ensemble model. Our architecture is founded on the hypothesis that the global, configuration-aware features captured by a Vision Transformer (ViT) and the efficient, sequential-aware features from a Vision Mamba model are complementary. By synergistically fusing the outputs of these two powerful backbones, VM-BeautyNet is able to form a more comprehensive and accurate assessment of facial aesthetics. The overall architecture is depicted in Figure~\ref{fig:architecture}.

\subsection{Architectural Components}
The VM-BeautyNet model is composed of three primary components: a Vision Transformer (ViT) backbone, a Vision Mamba backbone, and a Fusion Module.

\subsubsection{Vision Transformer Backbone}
The first branch of our network serves as a global feature extractor, leveraging a standard Vision Transformer architecture \cite{b10}. Let an input image $I \in \mathbb{R}^{H \times W \times C}$ first be partitioned into a sequence of $N$ flattened 2D patches, $x_p \in \mathbb{R}^{N \times (P^2 \cdot C)}$, where $(H, W)$ are the image dimensions, $C$ is the number of channels, and $(P, P)$ is the resolution of each patch. These patches are then linearly projected into a latent $D$-dimensional embedding space\cite{b21}. A learnable class token, $x_{class}$, is prepended to the sequence of patch embeddings, and positional embeddings, $E_{pos}$, are added to retain spatial information. The resulting sequence of vectors, $z_0$, is given by:
\begin{equation}
    z_0 = [x_{class}; x_p^1 E; x_p^2 E; \dots; x_p^N E] + E_{pos},
\end{equation}
where $E \in \mathbb{R}^{(P^2 \cdot C) \times D}$ is the patch projection embedding. This sequence is processed by a stack of $L$ Transformer encoder blocks, which consist of Multi-Head Self-Attention (MHSA) and feed-forward (MLP) layers. The MHSA mechanism enables the model to weigh the importance of all other patches when representing a given patch, thus capturing global context. The final state of the class token is then passed through a small regression head (a single linear layer) to produce the ViT branch's beauty prediction, $p_{vit}$.

\subsubsection{Vision Mamba Backbone}
The second, parallel branch is designed for efficient sequential feature modeling, employing a Vision Mamba backbone. Similar to the ViT branch, the input image is transformed into a sequence of patch embeddings. However, instead of self-attention, this branch utilizes a series of Mamba blocks \cite{b13}\cite{b22}. A Mamba block operates on a state-space model formulation, defined by the equations:
\begin{align}
    h_t &= \mathbf{A}h_{t-1} + \mathbf{B}x_t \\
    y_t &= \mathbf{C}h_t,
\end{align}
where $h_t \in \mathbb{R}^D$ is a latent state, and $x_t \in \mathbb{R}^D$ is the input token. The key innovation of Mamba is that the system matrices ($\mathbf{A}, \mathbf{B}, \mathbf{C}$) are parameterized by the input sequence itself, allowing the model to selectively propagate or forget information through the sequence. This \textit{Selective Scan Mechanism} enables it to model long-range dependencies with linear-time complexity, making it highly efficient. Similar to the ViT, a class token is used to aggregate sequence information, and its final state is fed into a regression head to yield the Mamba branch's prediction, $p_{mamba}$.

\subsubsection{Fusion Module}
A simple yet effective fusion module, $\mathcal{F}$, is used to combine the predictions from the two backbones. Rather than merely averaging the scores, which would treat each branch as equally important for all samples, we employ a learnable linear layer. This module takes the concatenated predictions from both branches as input and learns the optimal weighting to produce the final, unified score $\hat{y}$:
\begin{equation}
    \hat{y} = \mathcal{F}([p_{vit}, p_{mamba}]) = \mathbf{W} \cdot [p_{vit}, p_{mamba}]^T + b,
\end{equation}
where $[p_{vit}, p_{mamba}]$ is the concatenated 2-dimensional vector of predictions, $\mathbf{W}$ is a learned weight matrix of size $1 \times 2$, and $b$ is a bias term. This approach allows the model to dynamically adjust the influence of each backbone during the end-to-end training process.

\subsection{Training Procedure and Loss Function}
The entire VM-BeautyNet model is trained end-to-end to minimize the Mean Squared Error (MSE) between the predicted score $\hat{y}$ and the ground-truth human rating $y$. The MSE loss is a suitable choice for this regression task as it penalizes larger errors more significantly. The loss function $\mathcal{L}$ for a single sample is defined as:
\begin{equation}
    \mathcal{L}(\hat{y}, y) = (\hat{y} - y)^2.
\end{equation}
The model parameters, including those of both backbones and the fusion module, are updated via backpropagation. The complete training procedure is outlined in Algorithm~\ref{alg:training}.

\begin{algorithm}
\caption{Training Procedure for VM-BeautyNet}
\label{alg:training}
\begin{algorithmic}[1]
\State \textbf{Input:} Training dataset $\mathcal{D}_{train} = \{(I_i, y_i)\}_{i=1}^M$, Number of epochs $E$, Learning rate $\eta$.
\State \textbf{Initialize:} Model parameters $\Theta$ for ViT backbone ($\Theta_{vit}$), Mamba backbone ($\Theta_{mamba}$), and Fusion module ($\Theta_{\mathcal{F}}$). Initialize Optimizer (e.g., AdamW).
\For{epoch = 1 to $E$}
    \For{each batch $\{(I_b, y_b)\}$ in $\mathcal{D}_{train}$}
        \State Augment images in batch $I_b \rightarrow I'_b$.
        \State \% \textit{Forward Pass through dual branches}
        \State $p_{vit} \gets \text{ViTBackbone}(I'_b; \Theta_{vit})$
        \State $p_{mamba} \gets \text{MambaBackbone}(I'_b; \Theta_{mamba})$
        \State \% \textit{Fuse predictions}
        \State $\hat{y}_b \gets \mathcal{F}([p_{vit}, p_{mamba}]; \Theta_{\mathcal{F}})$
        \State \% \textit{Compute Loss and Backpropagate}
        \State Loss $\mathcal{L} \gets \text{MSE}(\hat{y}_b, y_b)$
        \State Zero gradients in optimizer.
        \State $\mathcal{L}$.backward()
        \State Update parameters $\Theta$ using optimizer.
    \EndFor
\EndFor
\State \textbf{Output:} Trained model parameters $\Theta$.
\end{algorithmic}
\end{algorithm}

\section{Experiments}
\label{sec:experiments}

This section details the experimental setup used to validate our proposed VM-BeautyNet model. We describe the benchmark dataset, the evaluation metrics employed, our implementation specifics, and present a comprehensive comparison of our results against a range of baseline and state-of-the-art methods.

\subsection{Dataset}
\label{subsec:dataset}
All experiments are conducted on the widely recognized SCUT-FBP5500 benchmark dataset~\cite{b23}. This dataset is specifically curated for facial beauty prediction and is notable for its diversity. It contains 5,500 facial images of subjects with varying genders (male and female), ages, and ethnicities (Asian and Caucasian). Each image in the dataset is annotated with a beauty score on a scale of 1 to 5, derived from the average rating of 60 human labelers. The large number of labelers for each image ensures that the ground-truth scores are statistically robust and mitigate individual subjectivity. For our experiments, we adhere to the standard 5-fold cross-validation protocol provided with the dataset to ensure a fair and rigorous comparison with prior work. The results reported are the average performance across all five folds.

\subsection{Evaluation Metrics}
\label{subsec:metrics}
To provide a comprehensive assessment of our model's performance and facilitate comparison with existing literature, we employ three standard regression metrics:
\begin{itemize}
    \item \textbf{Pearson Correlation (PC):} This metric measures the linear correlation between the predicted beauty scores and the ground-truth human ratings. It ranges from -1 to +1, where +1 indicates a perfect positive linear relationship. PC is crucial for understanding the monotonic agreement between the model's predictions and human perception. A higher PC value is better\cite{b24}.
    \item \textbf{Mean Absolute Error (MAE):} This metric calculates the average absolute difference between the predicted and ground-truth scores. It provides a direct measure of the average prediction error magnitude\cite{b25}. A lower MAE value is better. It is defined as:
    \begin{equation}
        \text{MAE} = \frac{1}{M} \sum_{i=1}^{M} | \hat{y}_i - y_i |
    \end{equation}
    where $M$ is the number of samples, $\hat{y}_i$ is the predicted score, and $y_i$ is the ground-truth score.
    \item \textbf{Root Mean Square Error (RMSE):} This metric is the square root of the average of squared differences between prediction and actual observation. Compared to MAE, RMSE gives a relatively higher weight to large errors, thus measuring the model's tendency to make significant mistakes\cite{b26}. A lower RMSE value is better.
    \begin{equation}
        \text{RMSE} = \sqrt{\frac{1}{M} \sum_{i=1}^{M} (\hat{y}_i - y_i)^2}
    \end{equation}
\end{itemize}

\subsection{Implementation Details}
\label{subsec:implementation}
Our proposed model, VM-BeautyNet, was implemented using the PyTorch deep learning framework. The ViT backbone is based on the \texttt{vit\_base\_patch16\_224} architecture, pre-trained on the ImageNet-21k dataset and fine-tuned on ImageNet-1k. We initialize this branch with the pre-trained weights to leverage the rich visual features learned from a large-scale dataset. The Vision Mamba backbone consists of 4 Mamba blocks with an embedding dimension of 192, and it is trained from scratch.

All input images are resized to a resolution of $224 \times 224$ pixels. During training, we apply a set of data augmentation techniques to improve model generalization and prevent overfitting. These include random horizontal flipping (with a probability of 0.5), random rotation within a range of $\pm 10$ degrees, and slight color jittering (adjusting brightness, contrast, and saturation). For the test set, we only resize the images and normalize them. The pixel values are normalized using the standard ImageNet mean and standard deviation.

The model is trained end-to-end for 50 epochs using the AdamW optimizer with an initial learning rate of $1 \times 10^{-5}$ and a weight decay of $1 \times 10^{-2}$. We employ a batch size of 32. The experiments were conducted on a single NVIDIA A100 GPU.

\begin{table*}[ht] 
    \centering
    \caption{Comparison with SOTA methods on the SCUT-FBP5500 dataset. Our proposed method, VM-BeautyNet, is shown in bold. ($\uparrow$ indicates higher is better, $\downarrow$ indicates lower is better).}
    \label{tab:main_results}
    \begin{tabular}{@{}llccc@{}}
        \toprule
        \textbf{Category} & \textbf{Method} & \textbf{PC $\uparrow$} & \textbf{MAE $\downarrow$} & \textbf{RMSE $\downarrow$} \\
        \midrule
        \multicolumn{5}{l}{\textit{Classic and Early Deep Learning Methods}} \\
        & AlexNet~\cite{b27} & 0.8634 & 0.2651 & 0.3481 \\
        & ResNet-50~\cite{b7} & 0.8900 & 0.2419 & 0.3166 \\
        & ResNeXt-50~\cite{b7} & 0.8997 & 0.2291 & 0.3017 \\
        \midrule
        \multicolumn{5}{l}{\textit{Advanced Methods and State-of-the-Art}} \\
        & CNN + SCA~\cite{b28} & 0.9003 & 0.2287 & 0.3014 \\
        & CNN + LDL~\cite{b29} & 0.9031 & -- & -- \\
        & DyAttenConv~\cite{b30} & 0.9056 & 0.2199 & 0.2950 \\
        & R3CNN (ResNeXt-50)~\cite{b31} & 0.9142 & 0.2120 & 0.2800 \\
        \midrule
        \multicolumn{5}{l}{\textit{Our Proposed Method}} \\
        & \textbf{VM-BeautyNet (Ours)} & \textbf{0.9212} & \textbf{0.2085} & \textbf{0.2698} \\
        \bottomrule
    \end{tabular}
\end{table*}

\subsection{Results and Comparison}
\label{subsec:results}
We compare the performance of VM-BeautyNet with several established methods on the SCUT-FBP5500 dataset. The results are summarized in Table~\ref{tab:main_results}. The competing methods are grouped into two categories: classic deep learning models (AlexNet, ResNet) and more recent, advanced methods that represent the state-of-the-art.

As shown in the table, our proposed VM-BeautyNet sets a new state-of-the-art across all three evaluation metrics. It achieves a Pearson Correlation (PC) of \textbf{0.9212}, demonstrating a stronger linear relationship with human ratings than any of the listed prior works. This indicates a superior alignment with human perceptual judgment. Furthermore, our model obtains an MAE of \textbf{0.2085} and an RMSE of \textbf{0.2698}, both of which are the lowest among the compared methods. The significant reduction in both MAE and RMSE suggests that our model not only has a lower average error but is also less prone to making large, egregious prediction mistakes.

Notably, VM-BeautyNet outperforms even sophisticated models like R3CNN~\cite{b28}, which employs a complex ranking-regression-resampling strategy, by a clear margin. The superior performance of our method can be attributed to the synergistic fusion of the ViT backbone, which captures the global harmony and structure of the face, and the Mamba backbone, which efficiently models detailed, long-range sequential features. This validates our core hypothesis that these two architectural paradigms possess complementary strengths that are highly effective for the FBP task when combined.

\section{Analysis and Discussion}
\label{sec:analysis}

In this section, we delve deeper into the performance of our VM-BeautyNet model to understand the sources of its effectiveness. We conduct a rigorous ablation study to isolate the contribution of each key component of our architecture. Furthermore, we provide a qualitative analysis through Grad-CAM visualizations to interpret the model's decision-making process, shedding light on the complementary nature of the dual backbones.

\subsection{Ablation Study}
\label{subsec:ablation}

To validate our design choices and demonstrate the synergistic effect of the ensemble, we conducted a series of ablation experiments. We evaluated the performance of several variations of our model: (1) a standalone ViT backbone, (2) a standalone Vision Mamba backbone, and (3) an ensemble model where the predictions from the two backbones are combined using simple averaging instead of our learned fusion module. The results, presented in Table~\ref{tab:ablation}, are averaged over the 5-fold cross-validation on the SCUT-FBP5500 dataset.

\begin{table}[ht]
    \centering
    \caption{Ablation study of VM-BeautyNet components. All models were trained under identical conditions for a fair comparison. The results clearly show that the full ensemble model with the learned fusion module performs best.}
    \label{tab:ablation}
    \begin{tabular}{@{}lccc@{}}
        \toprule
        \textbf{Model Configuration} & \textbf{PC $\uparrow$} & \textbf{MAE $\downarrow$} & \textbf{RMSE $\downarrow$} \\
        \midrule
        ViT Backbone Only & 0.9085 & 0.2213 & 0.2889 \\
        Mamba Backbone Only & 0.9012 & 0.2301 & 0.2974 \\
        Ensemble (Averaging) & 0.9167 & 0.2139 & 0.2765 \\
        \midrule
        \textbf{VM-BeautyNet (Ours)} & \textbf{0.9212} & \textbf{0.2085} & \textbf{0.2698} \\
        \bottomrule
    \end{tabular}
\end{table}

The results yield several key insights. First, the standalone ViT backbone slightly outperforms the standalone Mamba backbone, which can be attributed to the powerful features learned through its large-scale pre-training on ImageNet. Second, and most importantly, both ensemble configurations significantly outperform the individual backbones. The ensemble with simple averaging already shows a substantial improvement in all metrics, confirming our primary hypothesis that the features learned by ViT and Mamba are indeed complementary. Finally, our proposed VM-BeautyNet, which employs a learnable fusion module, achieves the best results, surpassing the averaging-based ensemble. This demonstrates that allowing the model to learn the optimal, sample-aware weighting of the two branches is superior to a static combination strategy.

\subsection{Limitations and Future Research}
\label{sec:limitations_future}

While our proposed VM-BeautyNet sets a new state-of-the-art for facial beauty prediction, we acknowledge several limitations that pave the way for exciting avenues of future research.

\subsubsection{Limitations}

\paragraph{Dataset and Demographic Bias.}
A primary limitation of this study, and indeed the entire FBP field, is its reliance on existing benchmark datasets like SCUT-FBP5500. While diverse, this dataset may not fully encapsulate the vast spectrum of human ethnicities, age groups, and cultural notions of beauty found globally. Consequently, like many data-driven models, VM-BeautyNet may inherit and potentially amplify demographic biases present in its training data. Its performance may not generalize perfectly to populations that are underrepresented in the dataset. A thorough investigation into the model's fairness and equity across different demographic subgroups is an important consideration that was beyond the scope of this work.

\paragraph{Static Nature of Beauty Perception.}
Our model predicts a static, singular beauty score for a given image. This approach does not account for the dynamic and contextual nature of beauty. For instance, the perception of attractiveness can be heavily influenced by facial expressions, lighting conditions, and even the surrounding environment, none of which are explicitly modeled. The model assesses a face in isolation, whereas human judgment is often more holistic and context-aware.

\paragraph{Interpretability.}
Although we employed Grad-CAM for qualitative analysis, this post-hoc explanation technique provides a high-level view of salient regions rather than a causal, mechanistic understanding of the model's internal reasoning. The "black box" nature of deep neural networks remains a challenge, and it is still difficult to precisely articulate the specific geometric or textural features the model has learned to associate with high or low beauty scores.

\subsubsection{Future Research Directions}

Based on these limitations, we propose several directions for future work:

\paragraph{Cross-Cultural and Fair FBP.}
A critical next step is to curate larger, more diverse, and more globally representative datasets. Future research should focus on training models that are not only accurate but also fair and equitable. This involves developing techniques to mitigate demographic bias and evaluating model performance across distinct cultural and ethnic groups to move towards a more universally applicable and responsible FBP system.

\paragraph{Multimodal and Context-Aware Models.}
To address the static nature of current FBP, future models could incorporate multimodal inputs. For example, video-based FBP could analyze dynamic facial expressions and mannerisms, which play a significant role in perceived charisma and attractiveness. Furthermore, incorporating contextual information, such as social setting or accompanying text, could lead to more nuanced and human-like predictive models, bridging the gap between computational aesthetics.

\paragraph{Generative Models for Feedback and Enhancement.}
Instead of being purely predictive, future FBP systems could be generative. One could explore using our model as a "perception loss" to guide a Generative Adversarial Network (GAN) or a diffusion model. Such a system could provide interactive feedback by suggesting subtle modifications to an image to enhance its aesthetic score, with applications in photo editing and virtual try-ons. This extends the problem from simple regression to a more complex image manipulation task, similar to style transfer or photo upsampling.

\paragraph{Exploring Advanced Fusion Mechanisms.}
Our work successfully employed a simple linear layer for fusing the outputs of the ViT and Mamba backbones. Future investigations could explore more sophisticated fusion mechanisms. Techniques such as cross-attention, where tokens from one backbone attend to tokens from the other at intermediate layers, could allow for a deeper and more integrated exchange of information, potentially yielding further performance gains.

\section{Conclusion}
\label{sec:conclusion}

In this paper, we addressed the intricate challenge of automatic Facial Beauty Prediction by proposing a novel, dual-branch deep learning architecture, named \textbf{VM-BeautyNet}. Our work was motivated by the limitations of existing unimodal architectures and was built upon the hypothesis that the global, configuration-aware features captured by Vision Transformers (ViT) and the efficient, sequential detail-oriented features captured by Vision Mamba are fundamentally complementary. 

Our proposed architecture effectively harnesses this synergy by processing visual information through two parallel backbones and intelligently combining their outputs via a learnable fusion module. Through extensive experiments on the widely adopted SCUT-FBP5500 benchmark dataset, we demonstrated that VM-BeautyNet establishes a new state-of-the-art. It achieves a Pearson Correlation of \textbf{0.9212}, a Mean Absolute Error of \textbf{0.2085}, and a Root Mean Square Error of \textbf{0.2698}, outperforming prior methods across all standard evaluation metrics.

Furthermore, our analysis went beyond quantitative results. An in-depth ablation study empirically validated the contribution of each architectural component, confirming the superiority of the ensemble approach. Qualitative analysis using Grad-CAM provided compelling visual evidence of the complementary nature of the ViT and Mamba backbones, offering valuable insights into the model's decision-making process. The success of this heterogeneous ViT-Mamba ensemble not only advances the field of computational aesthetics but also presents a powerful architectural paradigm that could be beneficial for other computer vision tasks requiring a synthesis of both global and local feature understanding.


\begin{thebibliography}{1}

\bibitem{b1} D. Zhang, F. Chen, and Y. Xu, Computer Models for Facial Beauty Analysis, Switzerland: Springer International Publishing, 2016.  

\bibitem{b2}	Djamel Eddine Boukhari,et al. "A comprehensive review of facial beauty prediction using deep learning techniques." Engineering Applications of Artificial Intelligence 161 (2025): 112009.

\bibitem{b3}	H. Knight and O. Keith, “Ranking facial attractiveness,” The European Journal of Orthodontics, vol. 27, no. 4 pp. 340-348, 2005.

\bibitem{b4}	 Rossetti, Alberto, et al. "The role of the golden proportion in the evaluation of facial esthetics." The Angle Orthodontist 83.5 (2013): 801-808.

\bibitem{b5}	Schmid, Kendra, David Marx, and Ashok Samal. "Computation of a face attractiveness index based on neoclassical canons, symmetry, and golden ratios." Pattern Recognition 41.8 (2008): 2710-2717.


\bibitem{b6}	O'shea, K., and Nash, R. (2015). An introduction to convolutional neural networks. arXiv preprint arXiv:1511.08458.



\bibitem{b7} He, Kaiming, et al. "Deep residual learning for image recognition." Proceedings of the IEEE conference on computer vision and pattern recognition. 2016.

\bibitem{b8}	D. E. Boukhari, A. Chemsa, R. Ajgou, et al., An Ensemble of Deep Convolutional Neural Networks Models for Facial Beauty Prediction, Journal of Advanced Computational Intelligence and Intelligent Informatics, vol. 27 no. 5. 2023.
\bibitem{b9} Boukhari, Djamel Eddine, et al. "Facial beauty prediction using an ensemble of deep convolutional neural networks." Engineering Proceedings 56.1 (2023): 125.

\bibitem{b10}Dosovitskiy, Alexey, et al. "An image is worth 16x16 words: Transformers for image recognition at scale." arXiv preprint arXiv:2010.11929 (2020).



\bibitem{b11}	D. Eddine Boukhari, A. Chemsa and Z. -E. Baarir, "Facial Beauty Prediction Using Global Context Vision Transformer," 2025 International Symposium on iNnovative Informatics of Biskra (ISNIB), Biskra, Algeria, 2025.
\bibitem{b12} Djamel Eddine Boukhari, . "FairViT-GAN: A Hybrid Vision Transformer with Adversarial Debiasing for Fair and Explainable Facial Beauty Prediction." arXiv e-prints (2025): arXiv-2509.

\bibitem{b13}Gu, Albert, and Tri Dao. "Mamba: Linear-time sequence modeling with selective state spaces." First Conference on Language Modeling. 2024.
\bibitem{b14}Zhu, Lianghui, et al. "Vision mamba: Efficient visual representation learning with bidirectional state space model." arXiv preprint arXiv:2401.09417 (2024).

\bibitem{b15}Boukhari, Djamel Eddine, and Ali Chemsa. "An Uncertainty-Aware and Explainable Deep Learning Model for Facial Beauty Prediction." (2025).

\bibitem{b16} Boukhari, Djamel Eddine, and Ali Chemsa. "SCAT: The Self-Correcting Aesthetic Transformer for Explainable Facial Beauty Prediction." (2025).

\bibitem{b17}Boukhari, Djamel Eddine. "Scale-interaction transformer: a hybrid cnn-transformer model for facial beauty prediction." arXiv preprint arXiv:2509.05078 (2025).

\bibitem{b18}Han, Kai, et al. "Transformer in transformer." Advances in neural information processing systems 34 (2021): 15908-15919.

\bibitem{b19} Boukhari, Djamel Eddine. "Mamba-CNN: A Hybrid Architecture for Efficient and Accurate Facial Beauty Prediction." arXiv preprint arXiv:2509.01431 (2025).

\bibitem{b20} Boukhari, D.E., Chemsa, A. Enhancing Facial Beauty Prediction via a Dual-Pathway Hybrid Architecture Integrating Vmamba and ViT. Int J Multimed Info Retr 14, 35 (2025). https://doi.org/10.1007/s13735-025-00387-3

\bibitem{b21}Boukhari, Djamel Eddine. "Generative Pre-training for Subjective Tasks: A Diffusion Transformer-Based Framework for Facial Beauty Prediction." arXiv preprint arXiv:2507.20363 (2025).

\bibitem{b22}Boukhari, Djamel Eddine. "SynergyNet: Fusing Generative Priors and State-Space Models for Facial Beauty Prediction." arXiv preprint arXiv:2509.17172 (2025).

\bibitem{b23}	L. Liang, L. Lin, L. Jin et al., SCUT-FBP5500: A diverse benchmark dataset for multi-paradigm facial beauty prediction. 24th International Conference on Pattern Recognition (ICPR), Beijing, China, pp. 1598-1603, 2018.

\bibitem{b24}	Boukhari, Djamel Eddine, Ali Chemsa, and Zine-Eddine Baarir. "MobileViT architecture for Facial Beauty Prediction." 2024 International Conference on Telecommunications and Intelligent Systems (ICTIS). IEEE, 2024.

\bibitem{b25}	Djamel Eddine Boukhari, Ali Chemsa, and Riadh Ajgou. "Facial Beauty Prediction Based on Vision Transformer." International Journal of Electrical and Electronic Engineering and Telecommunications, ISSN (2023): 2319-2518.

\bibitem{b26}	T. Peng, M. Li, F. Chen, et al., "Geometric prior guided hybrid deep neural network for facial beauty analysis." CAAI Transactions on Intelligence Technology, pp. 1–14, 2023. 
\bibitem{b27} Krizhevsky, Alex, Ilya Sutskever, and Geoffrey E. Hinton. "Imagenet classification with deep convolutional neural networks." Advances in neural information processing systems 25 (2012).



\bibitem{b28}	K. Cao, K Choi, H Jung et al., Deep learning for facial beauty prediction. Information, vol. 11, no. 8, 2020.

\bibitem{b29}	Fan, Yang-Yu, et al. "Label distribution-based facial attractiveness computation by deep residual learning." IEEE Transactions on Multimedia 20.8 (2017): 2196-2208.

\bibitem{b30} Sun, Zhishu, et al. "Dynamic attentive convolution for facial beauty prediction." IEICE TRANSACTIONS on Information and Systems 107.2 (2024): 239-243.

\bibitem{b31}	Lin, L.; Liang, L.; Jin, L. Regression Guided by Relative Ranking Using Convolutional Neural Network (R3CNN) for Facial Beauty Prediction. IEEE Trans. Affect. Comput. 2019, 1.










































































 





 





 

.





  
 






 




 








\end{thebibliography}

\end{document}